\documentclass[sigconf]{acmart}

\def\BibTeX{{\rm B\kern-.05em{\sc i\kern-.025em b}\kern-.08emT\kern-.1667em\lower.7ex\hbox{E}\kern-.125emX}}

\usepackage[ruled,vlined]{algorithm2e}
\usepackage{tabularx}

\copyrightyear{2023} 
\acmYear{2023} 
\setcopyright{acmlicensed}\acmConference[CIKM '23]{Proceedings of the 32nd ACM International Conference on Information and Knowledge Management}{October 21--25, 2023}{Birmingham, United Kingdom}
\acmBooktitle{Proceedings of the 32nd ACM International Conference on Information and Knowledge Management (CIKM '23), October 21--25, 2023, Birmingham, United Kingdom}
\acmPrice{15.00}
\acmDOI{10.1145/3583780.3615197}
\acmISBN{979-8-4007-0124-5/23/10}

\begin{document}

\title{POSPAN: Position-Constrained Span Masking for Language Model Pre-training}


\author{Zhenyu Zhang}
\authornote{The work has been accepted as CIKM21 short paper.}
\affiliation{%
  \institution{JD AI Research}
  \country{China}
}
\email{zhangzhenyu47@jd.com}

\author{Lei Shen}
\affiliation{%
  \institution{JD AI Research}
  \country{China}
}
\email{shenlei20@jd.com}

\author{Yuming Zhao}
\affiliation{%
  \institution{JD AI Research}
  \country{China}
}
\email{zhaoyuming3@jd.com}

\author{Meng Chen}
\affiliation{%
  \institution{JD AI Research}
  \country{China}
}
\email{chenmeng20@jd.com}

\author{Xiaodong He}
\affiliation{%
  \institution{JD AI Research}
  \country{China}
}
\email{xiaodong.he@jd.com}

\renewcommand{\shortauthors}{Zhenyu Zhang, Lei Shen, Yuming Zhao, Meng Chen, \& Xiaodong He}

\acmSubmissionID{2630}

\begin{abstract}
Span-level masked language modeling (MLM) has shown to be advantageous to pre-trained language models over the original single-token MLM, as entities/phrases and their dependencies are critical to language understanding. Previous works only consider span length with some discrete distributions, while the dependencies among spans are ignored, i.e., assuming that the positions of masked spans are uniformly distributed. In this paper, we present POSPAN, a general framework to allow diverse position-constrained span masking strategies via the combination of span length distribution and position constraint distribution, which unifies all existing span-level masking methods. To verify the effectiveness of POSPAN in pre-training, we evaluate it on the datasets from several NLU benchmarks. Experimental results indicate that the position constraint is capable of enhancing span-level masking broadly, and our best POSPAN setting consistently outperforms its span-length-only counterparts and vanilla MLM. We also conduct theoretical analysis for the position constraint in masked language models to shed light on the reason why POSPAN works well, demonstrating the rationality and necessity of POSPAN. 
\end{abstract}

\ccsdesc[500]{Computing methodologies~Lexical semantics}
\ccsdesc[500]{Computing methodologies~Natural language processing}

\keywords{Masked language modeling, pre-training, span length distribution, position constraint distribution}

\maketitle

\section{Introduction}
\label{sec:intro}
Large-scale pre-trained language models (PLMs) have achieved unmatched performance in many natural language understanding (NLU) tasks~\cite{bert, albert,xlnet, ernie2-aaai20}. 
As one of the most dominant model families, BERT-based models leverage masked language modeling (MLM)~\cite{bert}, a token-level denoised auto-encoding task, to facilitate the representation learning during pre-training. 
MLM samples tokens from the input text and replaces them with a special token ``[MASK]'', then asks the model to reconstruct original tokens based on the contextual hidden representations of masked tokens, enabling the model to capture semantics of each token with bi-directional context.
However, the vanilla MLM only considers masking individual tokens (words or sub-words) randomly, hence it neglects the information beyond token level, such as the compositional semantics for a phrase and the inter-token dependency for an entity, which are critical components of NLU in pre-training~\cite{ernie-ngram21}.

To learn the rich semantics in spans, e.g., phrases, entities, and n-grams, there has been an increasing body of works aiming to improve MLM~\cite{macbert20,ernie-ngram21, spanbert20, diabert-zhangcikm21}.
They mostly focus on span-level masking methods, where a contiguous token sequence, i.e., span, is masked according to some discrete span length distributions whose parameters are either preset~\cite{spanbert20} or adapted from external knowledge~\cite{ernie2-aaai20}.
Despite being advantageous over the vanilla MLM in \citet{spanbert20}, existing span-level masking methods all assume that the positions of masked spans are uniformly distributed, that is, the masked spans of a sequence are independent from each other~\cite{onloss20}.
This could cause sub-optimal selection of masked spans in pre-training, since the intrinsic features of natural languages, including semantic flows and inter-span interactions, can lead to strong dependencies of phrases and entities across context~\cite{macbert20}. 
Therefore, the assumption of uniformly-distributed positions of masked spans might not be a proper inductive bias, conversely, span masking bonded with certain position distance 
is likely to yield a better modeling of contextual dependencies.

To this end, we propose \textbf{POSPAN}, a general span masking algorithm that allows diverse \textbf{PO}sition-constrained \textbf{SPAN} masking strategies.
POSPAN employs a span length distribution $F_M$ and a position constraint distribution $F_D$ to control the length and position distance of masked spans, respectively. 
Moreover, all current span-level masking methods could be unified under POSPAN, where $F_M$s are different while $F_D$s are derived from the same uniform distribution.
To verify the effectiveness of POSPAN for language understanding, we evaluate PLMs with POSPAN on various tasks from the GLUE \cite{2019-GLUE} and Super GLUE \cite{wang2019superglue} benchmarks.
Experimental results show that POSPAN can outperform the previous span-length-only counterparts consistently when combined with suitable position constraints, which also matches the conclusion of our theoretical analysis. 

\section{Approach}
\label{sec:approach}

\subsection{Span Masking with Two Distributions}
\label{sec:ddr}

For span masking, there are two important factors: (1) the length of a span, i.e., how many consecutive tokens need to be masked, and (2) the position of a span, i.e., where to start masking.
Given a sequence with $m$ masked spans, $S=\{S_1, S_2, ..., S_m\}$, the length and position of the $i$-th span are denoted as $len_i$ and $pos_i$ (for simplicity, we use $i$ to represent the full notation of $i$-th span, $S_i$), we formulate these factors under two distributions:
\begin{align*}
&len_i \sim F_M, \\ 
&|pos_{i+1}-pos_i| \sim F_D, \,\,
0\leq pos_i < pos_{i+1} \leq N,
\end{align*}
where $F_M$ is the span length distribution, and $|pos_{i+1}-pos_i|$, named as \textbf{position constraint}, is the distance between two spans, $S_i$ and $S_{i+1}$, which follows the span constraint distribution $F_D$. $N$ is the length of input tokens.

Previous works mainly focus on the design of $F_M$, while the span position is randomly selected. 
We can derive that $F_D$ behind \textit{random selection} is a polynomial distribution. We start with two random spans, $S_i$ and $S_j$, and the probability of their position constraint with length $d$ is a new distribution $\hat{F_D}$:
\begin{equation}
\small
    P(|pos_i - pos_j| \leq d)= \hat{F_D}(d).
\end{equation}

\begin{algorithm}[t]
\small
\caption{Position-Constrained Span Masking}
\label{alg:masking}
\LinesNumbered 
\KwIn{$N$, $r_m$, $T$, $F_M$, $F_D$}
\KwOut{$T_{m}$}
{\footnotesize // masking rate: $r_m$, input tokens: $T$, input token length: $N$}\\
{\footnotesize // masked tokens: $T_{m}$}\\
{\footnotesize // initialize vectors: $m$, $d$, $T_m$}\\
$M$=vector($N$), $D$=vector($N$), $T_m$=copy($T$)\;
\For{i=1 $\to$ N}{
  sample span length $len_i \sim F_M$ \;
  sample position distance $d_i \sim F_D$ \;
  $M$[i]=$len_i$, $D$[i]=$d_i$\; 
}
spans=List(),
pos=1,
start=0,
do\_mask=False\;
\While{($pos < N$)}{
    start=pos \;
  \textbf{if}(do\_mask==False)  \textbf{then}  pos+=$D$[pos] \\
           \textbf{else}  pos+=$M$[pos] \;
  do\_mask=$\neg$do\_mask \;
  \textbf{if}(do\_mask==True)  spans.add(pair<start, pos-start>) \;
}
\textbf{while}(sum(spans.second) > $r_mN$) \,\,random.pop\_one(spans) \;
\For{span in spans}{
  set\_value($T_{m}$, span, [MASK]) \;
}
\end{algorithm}

Given the fixed masking rate $r_m \in (0,1)$, $r_m \times N$ tokens will be masked, and the position constraint $d$ is between 0 and $(1-r_m)N$, i.e., $0 \leq d\leq (1-r_m) N$. Then, $\hat{F_D}(d)$ can be obtained by integrating the position constraint over all positions:
\begin{equation}
\small
\begin{aligned}
\hat{F_D}(d)&=\frac{1}{c}
\bigg(\int_{0}^{d}(pos_j+d )\textrm{d}pos_j \\
&+ \int_{N-d}^{N} \Big(N-(pos_j-d )\Big)\textrm{d}pos_j \\
&+ \int_{d}^{N-d}\Big(pos_j+d -(pos_j-d)\Big)\textrm{d}pos_j \bigg) \\
&=\frac{2Nd-d^2}{c},
\end{aligned}
\end{equation}
where $c=(1-r_m^2)N^2$ is added to scale the result in range $(0,1)$.

Since there are $|S|$ spans uniformly sampled and the position constraint of each span pair follows $\hat{F_D}(d)$, the position constraint of two consecutive spans follows another polynomial distribution, denoted as $polyn(r_m, N)$:
\begin{equation}
\small
\begin{aligned}
polyn(r_m, N) \sim F_D &=P(pos_{i+1}-pos_i\leq \frac{d}{|S|-1}) \\
&\approx \hat{F_D}(d)
\end{aligned}
\end{equation}

It is clear that $F_D$ of existing methods is not fully considered, and we will explore the combination of different $F_M$s and $F_D$s in the following subsections.

\subsection{Theoretical Analysis of POSPAN}
\label{sec:app-theorem}

\subsubsection{Latent Semantic Dependency}

Assuming two spans of an input text are $S_{i}$ and $S_j$ whose start positions are $i$ and $j$, respectively. 
There are $len_i = |S_i|$ tokens in $S_i$ and $len_j = |S_j|$ tokens in $S_j$.
The contextual dependency of spans entails rich semantics that are important for the prediction of masked spans.
We use a latent variable $R_{ij}$ to represent the semantic dependency between $S_i$ and $S_j$.
In general, there are 3 cases for $R_{ij}$ in the natural language:
\begin{itemize}
\item \textbf{Case 1}: There are barely any dependency or semantic relationship between $S_i$ and $S_j$, i.e., we can predict $S_i$ and $S_j$ independently without knowing each other.
\item \textbf{Case 2}: $S_i \rightarrow S_{j}$, i.e., $S_i$ is the premise of $S_j$. When $S_i$ appears, $S_j$ will appear most of the time.
\item \textbf{Case 3}: $S_j \rightarrow S_{i}$, i.e., $S_j$ is the premise of $S_i$.
\end{itemize}

With the latent variable $R_{ij}$, we denote span masking methods as follows:
\begin{equation}
\small
\label{eq:app-train-all}
\begin{aligned}
&P(S_i, S_j|R_{ij}) = \frac{P(R_{ij}| S_i, S_j) * P(S_i, S_j)}{P(R_{ij})}, \\
&\text{log}P(S_i, S_j|R_{ij}) \propto \underbrace{\text{log}P(R_{ij}| S_i, S_j)}_{\textcircled{1}} + \underbrace{\text{log}P(S_i, S_j)}_{\textcircled{2}},
\end{aligned}
\end{equation}
where $P(R_{ij})$ is the prior probability that can be estimated from the corpus, and $P(S_i, S_j)$ is the span pair probability that is represented as $P(x_{i}, ..., x_{i+len_i-1}, x_{j}, ..., x_{j+len_j-1})$. 
Then, to maximize the log-likelihood of $M$ masked spans in training, we have:
\begin{equation}
\small
\label{eq:app-loss-span}
\begin{aligned} 
&\sum_{i,j}\text{log}P(S_i,S_j) = \frac{(M-1)\text{log}P(S_1, S_2,..., S_M)}{2} \\
&\propto \sum_{i=1}^{M}\text{log}P(S_i).
\\
&\mathcal{L}_{S}: \rightarrow max(\mathbb{E}[\text{log}P(S_i|len_i)]), \,\, where
\\
&\mathbb{E}[\text{log}P(S_i|len_i)] = \mathbb{E}_{len_{i} \sim F_M}(\sum_{l=0}^{len_i-1}\text{log}P(x_{i+l})).
\end{aligned}
\end{equation}

The assumption in Equation (\ref{eq:app-loss-span}) is that the probability of predicting a masked token is independent from each other. 
$\sum_{i,j}P(S_i,S_j)$ represents $M(M-1)/2$ combinations of $M$ unique spans.
We denote the objective function as the span loss $\mathcal{L}_{S}$, where span length $len_i \sim F_M$, and
``$:\rightarrow$'' means the goal of $\mathcal{L}_{S}$.

Previous works only focus on the second term in Equation (\ref{eq:app-train-all}) with various $F_M$s, and ignore the influence of latent semantic dependency $R_{ij}$ denoted as the first term that is governed by $F_D$.
For Case 1, the first term in Equation (\ref{eq:app-train-all}) is negligible. However, for Case 2 and 3, the first term in Equation (\ref{eq:app-train-all}) is critical, since improper settings of $F_D$ can harm span masking language models for natural language understanding.

\subsubsection{Position Constraint as Prior Knowledge}

The prediction of masked spans can be achieved via the usage of boundary tokens of a span \cite{spanbert20}, i.e., $P(S_i)$ can be estimated from the boundary tokens of $S_i$:
\begin{equation}
\small
\label{eq:app-sbo}
\begin{aligned}
P(S_i) &= P(x_{pos_i-1}, x_{pos_i + len_i}, pos_i)
\\
&\approx P(x_{pos_i-1}, x_{pos_i + len_i}, \hat{x_{pos_i}}), 
\end{aligned}
\end{equation}
where $pos_i$ is the position of $S_i$. 
Since we mask tokens in span $S_i$, the masked token $\hat{x_{pos_i}}$ is used to represent $pos_i$.
The distance dependency between $S_i$ and $S_j$ is reflected by $d$ tokens between $S_i$ and $S_j$, i.e., $I_{ij}=\{x_{pos_j - d}, ..., x_{pos_j-1}\}$.
Then, $R_{ij}$ is inferred by:
\begin{equation}
\small
\label{eq:fd-prob}
P(R_{ij}| d)=P(R_{ij}| S_i, S_j, I_{ij}).
\end{equation}

We assume that given $R_{ij}$, unmasked tokens in $I_{ij}$ are independent from tokens in $S_i$ and $S_j$, then the likelihood maximization of $P(R_{ij}|d)$ is equivalent to optimize the first term in Equation (\ref{eq:app-train-all}):
\begin{equation}
\small
    P(R_{ij}|S_i, S_j) \propto P(R_{ij}|d).
\end{equation}

Finally, the pre-training with masked language modeling can be decomposed into two losses:
\begin{equation}
\small
\begin{aligned} 
&\mathcal{L} = \mathcal{L}_{R} + \mathcal{L}_{S},
\\
&\mathcal{L}_{R}: \rightarrow max(\mathbb{E}[\text{log}P(R_{ij}|F_D)]),
\end{aligned}
\end{equation}
where $\mathcal{L}_{S}$ is span length loss from Equation (\ref{eq:app-loss-span}) and $\mathcal{L}_{R}$ is span dependency loss.
The span length and position constraint are controlled by the prior distributions $F_M$ and $F_D$, respectively, i.e., $len_i \sim F_M$ and $d \sim F_D$.
By properly setting the prior knowledge, we can improve the upper bound of pre-training for NLU tasks.

\subsection{POSPAN Algorithm}
\label{sec:algo}

Various masking strategies can be achieved via the combination of different $F_M$s and $F_D$s.
To investigate the impact of different masking strategies conveniently, we illustrate the sampling algorithm of POSPAN in Algorithm \ref{alg:masking}.
Given a text sequence, the algorithm first samples $N$ span lengths ($\sim F_M$) and $N$ inter-span position constraints ($\sim F_D$), stored in vector \textit{M} and \textit{D}, respectively (Line 3-8).
Then for each token position, we iteratively negate the value of \textit{do\_mask} to obtain the span length or position constraint turn by turn until all tokens are traversed, and select all possible spans into the set \textit{spans} (Line 9-15).
Next, we remove spans from \textit{spans} until the masked token number satisfies the masking rate requirement (Line 16).
Finally, we replace the selected tokens with ``[MASK]'' (Line 17-18).

We follow previous works~\cite{bert,roberta} to mask $15\%$ of tokens ($r_m=0.15$), where $80\%$ of them are replaced with ``[MASK]'', $10\%$ are replaced with tokens randomly sampled from the vocabulary, and the rest $10\%$ are untouched.
Through POSPAN, we can re-implement all the previous masking methods and design new masking strategies easily based on various distributions, including Normal ($Norm$), Geometric ($Geo$), Uniform ($Rand$) and Poisson ($Pois$) distribution. 
Table \ref{tab:distribution} illustrates the hyper-parameters of distributions we investigate, where the mean of these distributions are around 4 and 5 for $F_M$ and $F_D$, respectively, so as to be comparable with previous methods \cite{spanbert20, macbert20}.  
By combining different $F_M$s and $F_D$s, we develop several POSPAN settings to pre-train language models.

\begin{table}[htp]
\centering
\small
\begin{tabular}{lccc}
\hline 
\textbf{Notation} & \textbf{Distribution} & \textbf{$F_M$} & \textbf{$F_D$} \cr \hline
$Pois$ & Poisson & $\lambda=4$ & $\lambda=5$ \cr
$Norm$ & Normal & $\sigma$=1,$\mu$=4 & $\sigma$=1,$\mu$=5  \cr
$Geo$ & Geometric & $p$=0.2 & $p$=0.1 \cr
$Rand$ & Uniform & $a$=1,$b$=5 & $a$=4,$b$=6  \cr
\hline
\end{tabular}
\caption{Hyper-parameters of different distributions. We tune hyper-parameters of the distributions via grid search and find the best settings.}
\vspace{-6mm}
\label{tab:distribution}
\end{table}

\section{Experiments}
In this section, we first introduce the experimental setup. Then, we illustrate experimental results and conduct further discussions.

\begin{table*}[htp]
\centering
\small
\begin{tabular}{l|p{1.1cm}<{\centering}|p{1.8cm}<{\centering} p{0.6cm}<{\centering} p{0.6cm}<{\centering}|p{0.6cm}<{\centering} p{0.6cm}<{\centering}|p{1cm}<{\centering} p{1cm}<{\centering} p{0.6cm}<{\centering}}

\hline
Method & CoNLL & {MNLI(m/mm)} & MRPC & QNLI & BoolQ & COPA & ReCoRD & SQuAD & RACE \\ \hline
 {DeBERTaV3 \cite{he2021debertav3}} & {94.9} & {88.1/88.3} & {87.0} & {92.4} & {80.1} & {70.3} & {56.5/44.6} & {84.8/82.0} & {52.0} \\
 {MLM~\cite{bert}} & {95.3} & {88.2/88.5} & {88.4} & {92.5} & {80.5} & {70.9} & {56.3/44.9} & {84.8/82.1} & {52.1} \\
 \hline
 {Fixed} & {95.3} & {88.2/88.6} & {88.2} & {92.8} & {80.6} & {72.9} & {56.5/44.9} & {84.7/82.2} & {52.2} \\
 {N-gram~\cite{macbert20}} & {95.3} & {88.2/88.5} & {88.6} & {93.0} & {81.2} & {73.5} & {56.7/45.2} & {84.9/82.2} & {52.4} \\
 {WWM~\cite{bert-wwm}} & {95.2} & {88.2/88.5} & {88.0} & {92.7} & {80.8} & {71.8} & {56.4/44.7} & {84.8/82.2} & {52.3} \\
 {Geo~\cite{spanbert20}} & {95.7} & {88.5/88.7} & {88.9} & {93.1} & {81.3} & {73.2} & {56.8/45.1} & {85.0/82.5} & {52.5} \\
 {Pois~\cite{bart20}} & {95.6} & {88.4/88.7} & {87.5} & {93.0} & {81.0} & {73.9} & {56.7/45.1} & {85.1/82.5} & {52.3} \\
 \hline
 {POSPAN({WWM-$Norm$})} & {95.5} & {88.3/88.5} & {88.5} & {93.1} & {80.9} & {73.3} & {56.9/45.0} & {84.8/82.3} & {52.5} \\
 {POSPAN($Geo$-$Pois$)} & \textbf{95.9} & {88.8/89.0} & \textbf{89.2} & \textbf{93.4} & {81.6} & \textbf{75.7} & \textbf{57.3/45.6} & {85.4/82.5} & {52.8} \\
 {POSPAN($Pois$-$Pois$)} & {95.8} & \textbf{88.9/89.3} & {88.2} & {93.2} & \textbf{81.9} & {75.6} & {57.1/45.3} & \textbf{85.6/82.7} & \textbf{53.1} \\
 \hline
\end{tabular}
\caption{Experimental results of POSPAN. POSPAN($Geo$-$Pois$) denotes $F_M \sim Geo$ and $F_D \sim Pois$. CoNLL and SQuAD represent ConNLL 2003 and SQuAD v2.0. MNLI (m/mm) represents the two versions of MNLI, MNLI-matched and MNLI-mismatched.
}
\label{tab:exp-all}
\vspace{-3mm}
\end{table*}

\subsection{Experimental Settings}
\label{sec:expsettings}
\textbf{Datasets}.
We conduct experiments on four common types of NLU tasks, including named entity recognition (e.g., CoNLL 2003 \cite{conll2003}), sentence pair classification (e.g., MNLI \cite{williams2017broad-mnli}, MRPC \cite{dolan2005automatically-mrpc}, QNLI \cite{2019-GLUE}), question \& answering (e.g., BoolQ \cite{clark2019-boolq}, COPA \cite{roemmele2011choice-copa}), and machine reading comprehension (e.g., ReCoRD \cite{zhang2018-record} , SQuAD v2.0 \cite{2018-squad}, RACE \cite{lai-etal-2017-race}). For space limitation, we omit the details of each dataset. 
Considering the computational cost and experimental efficiency, we take the popular \textit{post-training} (i.e., the second-stage pre-training) strategy \cite{gururangan2020don, dialog-post-zhang-2023} with POSPAN instead of pre-training from scratch. We collect the text and remove labels from all training sets for post-training, which is about 1.5M sentences and 250M tokens. 
For \textit{fine-tuning}, all experiments were followed the setup in previous works \cite{roberta, he2021debertav3}.   


\noindent\textbf{Baselines}.
All experiments were conducted with the DeBERTaV3 \cite{he2021debertav3} backbone. The following baselines are compared:
(1) \textbf{DeBERTaV3} is the publicly available model checkpoint without post-training.
(2) \textbf{MLM} \cite{bert} post-trains DeBERTaV3 with sub-token masking, that is, the span length is 1.
(3) \textbf{Fixed} masks spans of length 4.
(4) \textbf{WWM} \cite{bert-wwm} masks spans of the whole word with several sub-tokens.
(5) \textbf{N-gram} \cite{macbert20} conducts masking where 10/20/30/40\% of spans are in length of 1/2/3/4.
(6) \textbf{Geo} \cite{spanbert20} and (7) \textbf{Pois} \cite{bart20}, in which the length of spans is sampled from Geometric and Poisson distribution, respectively. 
These methods use diverse strategies of span length masking ($F_M$s are different), but the same span position constraint ($F_D \sim polyn(r_m, N)$). For fair comparison, all models are trained with the same post-training and fine-tuning corpora and then evaluated on the same test sets as mentioned above. 

\noindent\textbf{Implementation Details}.
We load the publicly released checkpoint of of \textit{deberta-v3-xsmall} \cite{he2021debertav3} for model initialization. The number of hidden layers and attention heads is 12 and 6, and the hidden size, embedding size, and intermediate size is 384, 384, and 1536, respectively.
We first post-train a model for 20 epochs with batch size of 512, 5K warm-up steps, LAMB optimizer \cite{lamb-optimizer}, peak learning rate of 1e-4 with linear scheduler. 
Then we follow the setups in \citet{he2021debertav3, he2021deberta} to fine-tune the models with an extra classification or regression layer for each downstream dataset. Each model is fine-tuned for 10 epochs with batch size of \{16,32\} and learning rate of \{1e-5,2e-5\} with linear scheduler. We implement each model using Huggingface Transformers \cite{wolf-etal-2020-transformers}, report the average score of 10 runs on each dataset, and show the best result in bold with wilcoxon test ($p<0.05$) \cite{wilcoxon1945individual}.

\begin{figure}[t]
\begin{center}
\includegraphics[scale=0.31, trim={0 0 0 1cm}]{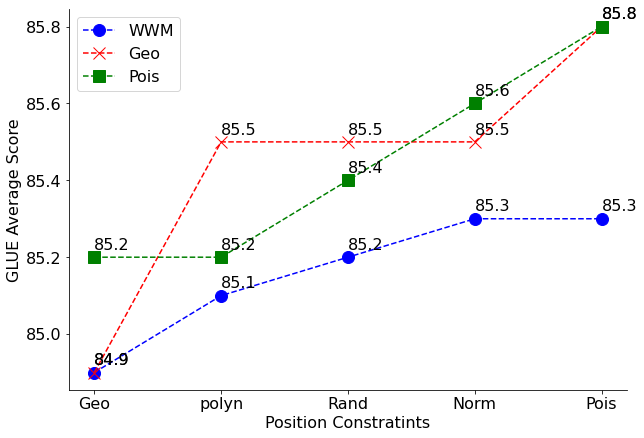}
\caption{The model performance of POSPAN with different position constraints ($x$-axis).}
\vspace{-3mm}
\label{fig:fd-dist}
\end{center}
\end{figure}

\subsection{Main Results}
\label{app:main-exp}

Table \ref{tab:exp-all} illustrates the experimental results. It's observed that: (1) All post-training models can bring further improvements compared to the strong baseline DeBERTaV3, which shows the effectiveness of post-training. (2) Compared with single-token masking, all span-level masking methods yield substantial improvements, which indicates the advantage of span-level masking on capturing the critical semantics of language. (3) Our proposed POSPAN obtained the best performance across different tasks. Specially, POSPAN surpassed the previous best baselines by 1.8\%, 0.9\%, 0.6\% on COPA, BoolQ, RACE respectively. It demonstrates the superiority and necessity of position constraint for span masking.

Previous research \cite{probagen-noahark20} has shown that the masking probability can introduce a type of prior knowledge for language models. We conjecture that POSPAN also introduces $F_M$ and $F_D$ as two kinds of prior knowledge, where $F_M$ can help the model capture n-gram-level sub-structures \cite{spanbert20, macbert20,ernie-ngram21} and $F_D$ is capable of catching the semantic dependencies among spans. Besides, the theoretical analysis in Section \ref{sec:app-theorem} also proves the necessity of both span length distribution and position constraint distribution for language model pre-training, which explains why POSPAN works. 

\subsection{Discussions}

To investigate the position constraint in POSPAN, we combine WWM, Geo, and Pois with various $F_D$s. 
Figure \ref{fig:fd-dist} shows the model performance under different $F_D$s on the GLUE benchmark \cite{2019-GLUE}, and ``$polyn$'' represents the span-length-only counterparts.
It shows that: 
(1) Position constraint distribution has a prominent impact on the model performance. 
For example, POSPAN($Pois$-$Pois$) improves the Pois by 0.6\%, while POSPAN($Geo$-$Geo$) decreases the score of Geo by 0.6\%.
(2) POSPAN with $F_D \sim Geo$ consistently harms the original span masking methods, while other $F_D$ distributions boost the performance of original span masking by different extent. 

The above results corroborate our hypothesis, that is, span positions are indeed not independent from each other, and utilizing $F_D$ to represent dependent span positions and model contextual dependency is beneficial to downstream tasks universally.
Mathematically speaking, when sampling span positions randomly, the distribution $polyn(r_m, N)$ promotes small inter-span distances, which prevents span dependency modeling, thus hinders the learning of contextual semantics.
Similarly, $F_D \sim Geo$ also promotes small inter-span distances and results in unsatisfying performance.
Empirically, the discrete distribution $Pois$ performs relatively better than other continuous distributions.
Such performance bias caused by distributions might indicate the discrete and flexible length of dependency in natural language.


\section{Conclusion and Future Work}
In this paper, we propose POSPAN, a novel position-constrained span masking method for language model pre-training. POSPAN leverages span length and position constraint distributions to mask tokens, and works as a general framework to unify existing span-level masking methods.
Extensive experiments are conducted to verify the effectiveness of POSPAN on various NLU tasks. 
Moreover, the theoretical analysis reveals the rationality and necessity of both span length distribution and position constraint distribution, which encourages language models to learn span-level semantics and their contextual dependencies. For the future work, we will explore more effective masking strategies by designing better $F_M$s and $F_D$s.

\bibliographystyle{ACM-Reference-Format}
\balance 
\bibliography{ref}


\end{document}